\pdfoutput=1

\documentclass[11pt]{article}

\usepackage[final]{acl}
\usepackage{enumitem}
\usepackage{times}
\usepackage{latexsym}
\usepackage{booktabs}
\usepackage{caption}
\usepackage{subcaption}
\usepackage[T1]{fontenc}

\usepackage[utf8]{inputenc}

\usepackage{microtype}

\usepackage{inconsolata}

\usepackage{graphicx}
\usepackage{multirow}

\title{Impact of Decoding Methods on Human Alignment of Conversational LLMs}

\author{
 \textbf{Shaz Furniturewala\textsuperscript{1,3}\thanks{Work done during internship at NUS Center for Trusted Internet and Community}},
 \textbf{Kokil Jaidka\textsuperscript{2,3}},
 \textbf{Yashvardhan Sharma\textsuperscript{1}},
\\
 \textsuperscript{1}Birla Institute of Technology and Science, Pilani \\
 \textsuperscript{2}NUS Center for Trusted Internet and Community, National University of Singapore
\\
 \textsuperscript{3}Department of Communications and New Media, National University of Singapore
}

\begin{document}
\maketitle
\begin{abstract}
To be included into chatbot systems, Large language models (LLMs) must be aligned with human conversational conventions. However, being trained mainly on web-scraped data gives existing LLMs a voice closer to informational text than actual human speech. In this paper, we examine the effect of decoding methods on the alignment between LLM-generated and human conversations, including Beam Search, Top K Sampling, and Nucleus Sampling. We present new measures of alignment in substance, style, and psychometric orientation, and experiment with two conversation datasets. Our results provide subtle insights: better alignment is attributed to fewer beams in Beam Search and lower values of P in Nucleus Sampling. We also find that task-oriented and open-ended datasets perform differently in terms of alignment, indicating the significance of taking into account the context of the interaction. 
\end{abstract}

\section{Introduction}
As large language models (LLMs) continue to evolve, their integration into chatbot systems has increasingly focused on not just understanding but also on aligning with human conversational norms. 
Models are trained and finetuned to be 'perfect assistants' which has inadvertently given them a voice that is eager, overly enthusiastic, and marked by use of words and phrases that feature prominently in informational and instructional texts but not so much in true human conversations~\cite{zhou2024real}. Therefore, LLM-human alignment is a crucial problem and has been studied across various contexts, such as coding, problem-solving, summarization, translation, and reasoning~\cite[for a review, see][]{shi2024thorough}. Among various techniques explored to improve this alignment, the perturbation of decoding parameters—such as Beam Search, Top K Sampling, and Nucleus Sampling—has shown promise. These decoding methods, encompassing both deterministic strategies like beam search and stochastic approaches such as temperature scaling, fundamentally influence how a model generates text. Preliminary studies suggest that while deterministic methods may better adhere to specific instructions, stochastic methods like P and K sampling could excel in scenarios involving unaligned models by introducing variability in responses~\cite{shi2024thorough}. Despite their potential, the impact of these methods on the quality of chatbot outputs, particularly in mimicking human conversational patterns, has not been comprehensively analyzed. \\
Achieving a high degree of alignment between the outputs of these models and actual human interactions is crucial not only for maintaining the natural flow of dialogue but also for ensuring the relevance and contextuality of the responses provided by chatbots. Yet, current evaluation methods are limited in their ability to assess whether these systems successfully emulate the human-like attributes essential for nuanced interactions. For instance, most work focuses on automatic evaluation methods such as BLEU, ROUGE, and METEOR with some others using classifiers trained on human judgement \cite{metrics}. While there has been some work in the creation of psychological metrics \cite{psychmetrics}, it merely focuses on broad aspects of dialog like emotion and personality. A study of dialogue dynamics requires an understanding of the deeper subtleties of interpersonal engagement beyond content, such as style and psychological orientation. Unlike emotion, style and psychological orientation are nuanced and multifaceted aspects of communication that have not been studied as much and are harder to accurately measure and control. \\
This paper aims to bridge this gap by systematically investigating the effects of different decoding methods on the alignment between chatbot outputs and human-like responses. We hypothesize that adjusting these decoding parameters can significantly enhance the naturalistic appeal and user engagement of chatbot conversations. To test this hypothesis, we employ a novel methodological approach, analyzing the performance of conversational LLMs through a series of experiments involving real human conversations. Our work offers the following contributions:
\begin{itemize}[noitemsep]
\item Two new parallel corpora of synthetic LLM-generated conversations, curated through turn-by-turn prompts with real-world dialogues sourced from two human-human datasets, collected across a variety of decoding methods. 
    \item New metrics for measuring LLM alignment to human conversations in substance, style, and psychometric orientation. 
\end{itemize}
Our findings aim to provide deeper insights into the practical applications of decoding methods and their potential to improve the human-likeness of chatbot interactions, thereby guiding future developments in chatbot design and deployment.

\section{Empirical Evaluation}
In this section we describe the datasets used for our experiments, the metrics employed to measure humane conversational traits, and the decoding methods used in the LLM's generation process. We created a turn-by-turn synthetic dataset of LLM generated conversations, adhering to a structured process across each conversation turn. Each conversation began with the opening turns of a conversation from one of the two datasets we considered—BOLT and CraiglistBargains—and we invited each LLM we considered to generate the next utterance by the speaker indicated. We then evaluated human-LLM conversation alignment along dimensions of Style, Psychometrics, and Semantic content. In this work we use Llama 3 (8B) and Llama 3 Instruct (8B) for our experiments.  Further, we vary the decoding methods during generation utilising Beam Search, Top K Sampling, and Nucleus Sampling, to gain insights into their impact on the quality of generated conversations. 
\subsection{Datasets}
\textbf{BOLT SMS/Chat Dataset}
\cite{bolt}, developed by the Linguistic Data Consortium consists of naturally occurring English conversations involving native speakers.  The corpus contains 18,429 two-person conversations totaling 3,674,802 words across 375,967 messages. For the purposes of this work, 2640 conversations ranging from 5 to 125 turns were used. \\
\textbf{CraigslistBargains}
\cite{he-etal-2018-decoupling} is a collection of 6682 human-human negotiation conversations between AMT agents. The agents are assigned the role of buyer and seller and are asked to negotiate the price of a real Craigslist listing. For this work, 5357 conversations ranging from 5 to 28 turns were used.

\subsection{Measures}
The following 6 metrics measure the stylistic, psychometric, and semantic similarity between the human and LLM generated texts. They are relative measures, using the original text as a reference in comparison with the LLM generated text. Each measure is computed at the utterance level and averaged across the entire conversation to arrive at a score. 

\subsubsection{Stylistic}
Style is a broad concept with various aspects. For the purposes of this paper we picked two aspects that are relevant to the datasets being used and are significantly impacted by the decoding parameters as seen in Figure \ref{fig:coeff_graph}. \\
\textbf{Politeness}
We used the ConvoKit Library \cite{convokit} to compute 21 characteristics representing facets of politeness, including deference, hedging, gratitude, factuality, among others. We then calculated the cross entropy score between each human and LLM generated utterance. \\
\textbf{Negotiation}
Based on the work done by \citet{harbingers}, we extracted 8 linguistic cues from each utterance including Claim, Premise, Contingency, Expansion, Temporal (Past and Future), Subject, and Comparison. We use these linguistic cues as a negotiation vector and compute the cross entropy score between the human and LLM generated utterance.

\subsubsection{Pyschometric}
\textbf{Self Concept}
We annotated 10,956 text messages from the BOLT dataset for the presence of three characteristics of self concept: Autonomy, Competence, and Relatedness. These annotations were done by Amazon Mechanical Turk workers on an interface we designed that provided positive and negative examples of each characteristic. We finetuned a classifier on this data and computed the cross entropy score between the predictions for the human and LLM generated utterances. \\
\textbf{Empathy}
We finetune an empathy classifier on the dataset created by \citet{empathy}. It contains 1860 short texts annotated for empathic concern. This classifier predicts the presence of empathic concern in the human and LLM generated utterances and we compute the cross entropy score between them. 

\begin{figure*}[h]
    \centering
    \includegraphics[width=\textwidth]{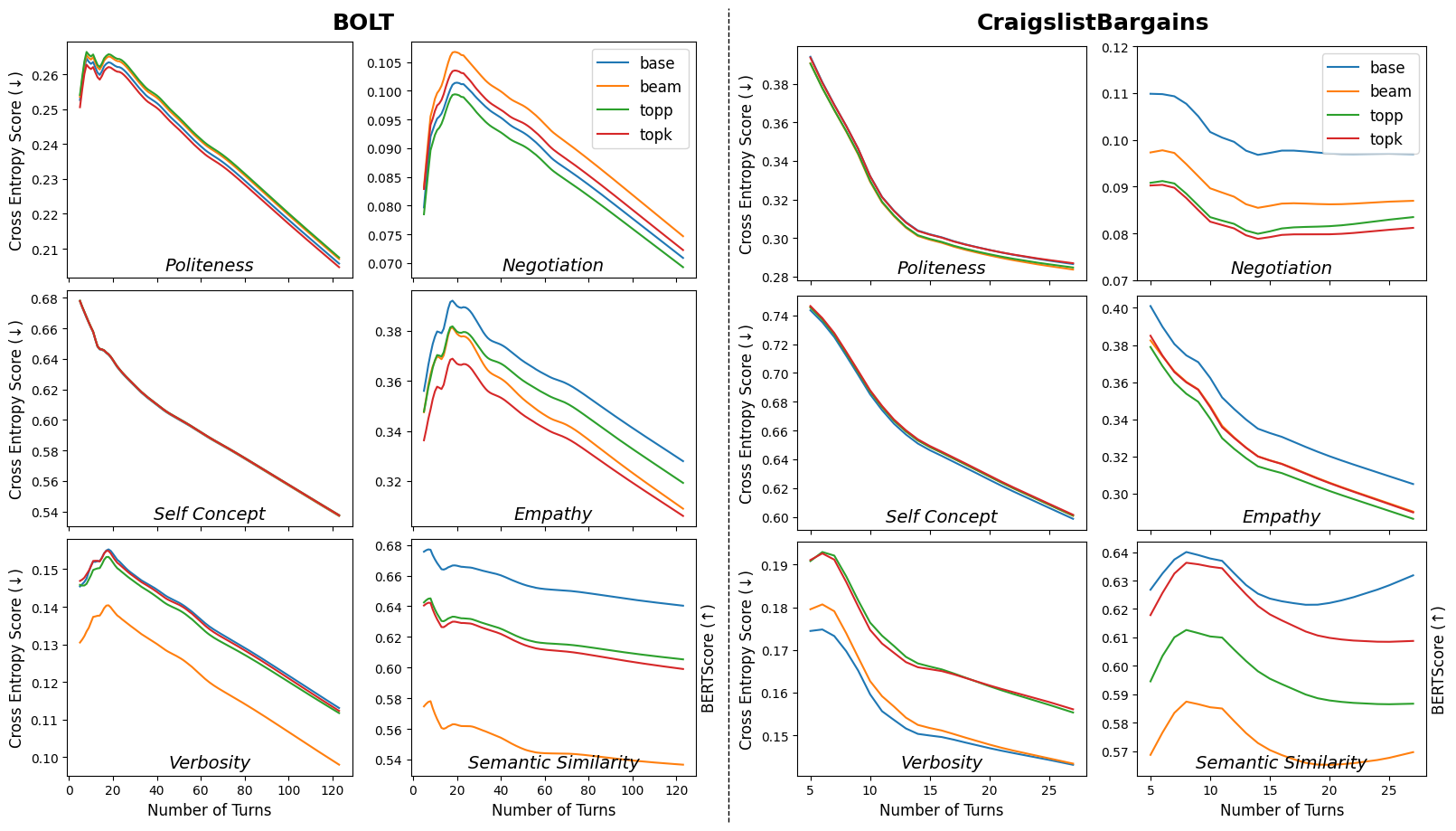}
    \caption{Turn-based scores for each decoding parameter, averaged across all perturbations and both models (Llama 2 and Llama 3).}
    \label{fig:comp_graph}
\end{figure*}

\subsubsection{Semantic}
\textbf{Verbosity}
For each utterance, we measure verbosity as the absolute difference between the length of the human and LLM generated utterances. \\
\textbf{Semantic Similarity}
We compute the semantic similarity between the human and LLM generated utterances using BERTScore \cite{bert-score}.  
\subsection{Decoding Parameters}
The standard generation setup uses the default temperature value of 1.0 and deterministic greedy decoding with no sampling. In the appendix we also display results on temperature perturbations. \\
\textbf{Beam Search} \cite{luong-etal-2015-addressing, graves2012sequence}
Using this decoding strategy, we can allow the model to evaluate multiple hypotheses at a time and ultimately pick the sequence that has the highest overall probability. While it is computationally more expensive, it can generate sequences that begin with low probability tokens but have the overall highest probability. In this work we evaluate beam search with 2, 4, 6, and 8 beams. \\
\textbf{Top K Sampling}
Introduced by \citet{topk}, this generation strategy filters out the K most probable next tokens and redistributes the probability mass among them. Then, based on their new probabilities, the next token is randomly chosen among them. In this work we evaluate Top-K Sampling with K = 30, 40, 50, 60, and 70. \\
\textbf{Nucleus Sampling (Top P)} \cite{holtzman2020curiouscaseneuraltext}
This sampling method filters the smallest number of tokens whose probability cumulatively exceeds P. In this manner, it dynamically changes the number of tokens being filtered based on the probability distribution. We evaluate Nucleus Sampling with P = 0.6, 0.7, 0.8, 0.9, 1.0.



\begin{table}
\small
\centering
\begin{tabular}{cc|cc|cc}   
\toprule
Beams & Change & P & Change & K & Change \\ \midrule
\multirow{2}{*}{2} & \multirow{2}{*}{\textbf{3.82\%}} & 0 & 2.35\% & 1 & 1.64\% \\
 & & 0.5 & -1.80\% & 20 & -3.21\% \\
\multirow{2}{*}{4} & \multirow{2}{*}{1.37\%}& 0.6 & 1.85\% & 30 & -5.02\% \\
 & & 0.7 & \textbf{2.50\%} & 40 & -1.19\% \\
\multirow{2}{*}{6} &  \multirow{2}{*}{-2.33\%}& 0.8 & -5.66\% & 50 & -1.33\% \\
 &  & 0.9 & -3.09\% & 60 & -0.55\% \\
8 &  0.21\%& 1 & -2.67\% & 70 & \textbf{2.99\%} \\
\bottomrule
\end{tabular}

\caption{Average change in alignment across the six metrics for various values of the decoding methods.}
\label{tab:imp_table}
\end{table}

 
\section{Results}
In this section we will analyse the results of the experiments described in Section 2. Initially, our analysis focuses on identifying how variants of different decoding parameters perform, then we examine turn-level results. \\
Table \ref{tab:imp_table} displays the average change in alignment scores over the base decoding method for each decoding parameter perturbation. The change in comparison to base is measured as the decrease in the cross entropy score of the generated text with the ground truth. Thus, a higher percentage change represents a greater decrease in the cross entropy score indicated better alignment with the human responses. These scores are averaged across both datasets (BOLT, CraigslistBargains) and across both models (Llama 3, Llama 3 Instruct). \\
We notice that using 2 Beams outperforms the base greedy decoding strategy, however, further increasing the number of beams diminishes this increase in performance, indicating a potential local minimum (or a local maximum in alignment). Lower values of P (0.6-0.7) have the best performance while P=1.0 demonstrates a significant decrease in alignment compared to base (greedy) decoding. This indicates that some of the least likely tokens in the vocabulary contribute to the drop in alignment when included in the sequence. 
Finally, there no observable trend in the perturbations of Top K Sampling with all values of K performing, on average, similarly to the base method, i.e., greedy decoding. \\ 
In Figure \ref{fig:comp_graph}, we plot the scores (scaled down and smoothened) divided by dataset and along with the number of turns in the conversation. This allows us to examine task-specific performance as a function of the length of the conversation. We see that as conversations get longer, the LLM is able to more accurately emulate these traits. Notably, this trend applies for negotiation on BOLT but not on CraigslistBargains where the performance quickly plateaus. For both datasets, Politeness, Self-Concept, Empathy, and Verbosity follow a similar trend of improving performance as the conversations get longer with beam search and nucleus sampling consistently outperforming Top K sampling. This is consistent with our previous conclusions about these two decoding methods. In addition, it indicates that for these metrics, alignment is correlated with the amount of context provided. 
\begin{figure}[h]
    \centering
    \includegraphics[width=0.45\textwidth]{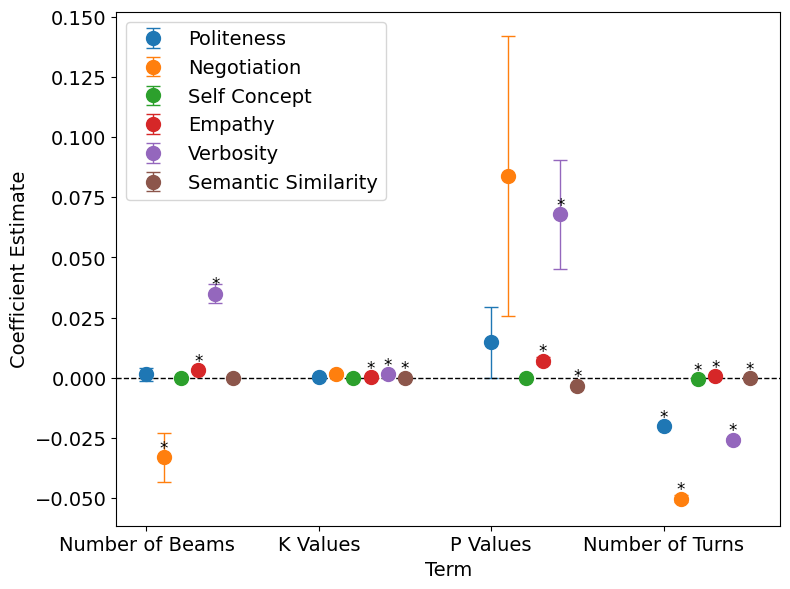}
    \caption{The parameters effecting significant positive and negative changes in style, psychometrics and semantics of LLM conversations. Calculated using multi-level models controlling for model and dataset differences.}
    \label{fig:coeff_graph}
\end{figure} \\
To validate our insights and conclusions from these experiments, we compute the correlation coefficients of the decoding parameters with the measures. In Figure \ref{fig:coeff_graph} we plot the coefficients of multilevel linear models for each of the six metrics calculated for the three decoding methods—Beam Search, Top K Sampling, Top P Sampling—and the number of turns or the length of a conversation. A positive coefficient indicates that a high value for that parameter predicts better alignment and a negative coefficient predicts worse alignment. Asterisks represent statistically significant associations (p$<$0.001). The first notable observation from the figure is that Top K sampling does not have any significant non-zero coefficients for any of the given metrics. Beam search only has non-zero coefficients with Negotation and Verbosity, having a positive coefficient for the former and negative one for the latter. Politeness, Negotiation, and Verbosity all have positive coefficients for P value and proportionally negative correlations for the number of turns. 
\section{Discussion and Conclusion}
The broader context of the datasets appears to affect the quality of generated conversations, as BOLT, being a chit-chat dataset, does not require the same amount of negotiation as the task oriented CraigslistBargains and allows the LLM to adapt to these traits quickly. Similarly, the models show decreasing Semantic Similarity on BOLT compared to CraigslistBargains where performance stays consistently high across conversations. The goal-oriented task of CraigslistBargains tends to have highly probable responses in a specific direction for each input. On the contrary, BOLT is very open-ended with each dialog allowing the conversation to go in many different directions. A similar effect is seen in the quality metrics, where over the course of a long conversation, the lack of structure in the task is seen to lead to more deviations by the LLM in BOLT compared to CraigslistBargains, manifesting as the decreased alignment performance seen in the graph. \\
Our experiments suggest that lower P values improve instruction adherence, while top-K sampling, unlike nucleus sampling, has a smaller impact, as the fixed number of tokens being filtered each time results in much less control over the redistributed probability mass as compared to variable tokens with a fixed cumulative probability threshold.  Thus, the best decoding method for human aligned conversational LLM output is likely a combination of \textbf{Low P Nucleus Sampling and Beam Search with  a small number of beams.} \\
A larger number of beams incorporates more obscure, lower probability words into the sequence that leads to worse alignment, possibly through introducing linguistic artifacts such as obscure words and longer texts to the sequence that undo the potential advantages of having more beams. 

\section{Limitations}
One particular limitation of our work is the usage of two specific aspects of style: Politeness and Negotiation. To ensure concise insights we limited the experiments to these two aspects since they pertain the most to the task specific dataset we used (CraigslistBargains). We believe the results observed for these two aspects should translate to other facets of style on other datasets and we hope to address this by expanding on these experiments in our future work.

\section{Acknowledgements}
This work is supported by the Ministry of Education, Singapore under its MOE AcRF TIER3 Grant (MOE-MOET32022-0001). The travel grant for this research is supported by the Department of Communication and New Media at the National University of Singapore.

\bibliography{custom}

\appendix

\section{Appendix}
\label{sec:appendix}
\subsection{Classifiers}
For the metrics computing Empathy and Self Concept we trained classifiers on annotated data to predict the presence of these attributes in the text. Both classifiers were finetuned variants of Microsoft's DeBERTa V3 Large model with 304 M parameters. \\
For empathy, this classifier was trained on the NewsEmpathy \cite{empathy} dataset containing 1860 instances of annotated text. The model was finetuned for 4 epochs, with a learning rate of 5e-6 and weight decay of 0.01. It achieved a validation F1 Score of 0.71.
For Self Concept, the classifier was trained on a dataset we annotated. It consists of 10956 texts extracted from the BOLT dataset and annotated for the presence of Autonomy, Competence, and Relatedness. This model was trained for 1 epoch with a learning rate of 1e-5 and weight decay of 0.01. It achieved a validation macro F1 score of 0.83.
\subsection{Impact of Instruction Finetuning}
\begin{table*}[]
\small
\centering
\begin{tabular}{c|ccccc|c}
\toprule
\multicolumn{1}{l}{} & Politeness                   & Negotiation                  & Self Concept                 & Empathy                      & Verbosity                    & Semantic Similarity          \\ \midrule
Llama 3 Instruct     & 0.312                        & 0.099                        & 0.666                        & 0.541                        & 0.232                        & 0.622                        \\ \midrule
\multicolumn{7}{c}{Number of Beams}                                                                                                                                                                            \\ \midrule
2                    & 0.311                        & 0.093                        & 0.666                        & 0.546                        & 0.221                        & \textbf{0.639}               \\
4                    & {\color[HTML]{C00000} 0.314} & 0.087                        & {\color[HTML]{C00000} 0.667} & 0.544                        & \textbf{0.216}               & 0.628                        \\
6                    & 0.312                        & \textbf{0.073}               & 0.667                        & {\color[HTML]{C00000} 0.548} & 0.227                        & {\color[HTML]{C00000} 0.619} \\
8                    & \textbf{0.310}               & {\color[HTML]{C00000} 0.104} & \textbf{0.666}               & \textbf{0.530}               & {\color[HTML]{C00000} 0.231} & 0.629                        \\ \midrule
\multicolumn{7}{c}{P Value}                                                                                                                                                                                    \\ \midrule
0                    & 0.315                        & 0.104                        & {\color[HTML]{C00000} 0.667} & 0.558                        & \textbf{0.203}               & 0.640                        \\
0.5                  & 0.315                        & 0.101                        & 0.667                        & 0.544                        & 0.213                        & \textbf{0.646}               \\
0.6                  & 0.313                        & {\color[HTML]{C00000} 0.111} & 0.667                        & 0.568                        & 0.207                        & 0.606                        \\
0.7                  & {\color[HTML]{C00000} 0.315} & 0.108                        & 0.666                        & 0.543                        & 0.214                        & 0.620                        \\
0.8                  & \textbf{0.312}               & 0.104                        & 0.666                        & \textbf{0.516}               & {\color[HTML]{C00000} 0.229} & 0.612                        \\
0.9                  & 0.313                        & \textbf{0.095}               & \textbf{0.665}               & 0.538                        & 0.218                        & {\color[HTML]{C00000} 0.604} \\
1                    & 0.313                        & 0.096                        & 0.666                        & {\color[HTML]{C00000} 0.570} & 0.220                        & 0.644                        \\ \midrule
\multicolumn{7}{c}{K Value}                                                                                                                                                                                    \\ \midrule
1                    & {\color[HTML]{C00000} 0.315} & 0.103                        & 0.667                        & 0.558                        & \textbf{0.203}               & \textbf{0.640}               \\
20                   & 0.313                        & {\color[HTML]{C00000} 0.120} & {\color[HTML]{C00000} 0.668} & 0.534                        & 0.214                        & {\color[HTML]{C00000} 0.615} \\
30                   & 0.307                        & 0.110                        & 0.666                        & 0.538                        & 0.221                        & 0.620                        \\
40                   & \textbf{0.306}               & 0.108                        & 0.667                        & \textbf{0.504}               & 0.225                        & 0.622                        \\
50                   & 0.309                        & 0.106                        & \textbf{0.665}               & 0.536                        & {\color[HTML]{C00000} 0.240} & 0.635                        \\
60                   & 0.312                        & \textbf{0.088}               & 0.666                        & {\color[HTML]{C00000} 0.560} & 0.212                        & 0.627                        \\
70                   & 0.312                        & 0.103                        & 0.666                        & 0.545                        & 0.224                        & 0.629                        \\ \midrule
\multicolumn{1}{l}{} & \multicolumn{1}{l}{}         & \multicolumn{1}{l}{}         & \multicolumn{1}{l}{}         & \multicolumn{1}{l}{}         & \multicolumn{1}{l}{}         & \multicolumn{1}{l}{}         \\ \midrule
Llama 3              & 0.316                        & 0.102                        & 0.661                        & 0.661                        & 0.174                        & 0.553                        \\ \midrule
\multicolumn{7}{c}{Number of Beams}                                                                                                                                                                            \\ \midrule
2                    & {\color[HTML]{C00000} 0.315} & \textbf{0.082}               & 0.642                        & 0.666                        & \textbf{0.141}               & {\color[HTML]{C00000} 0.465} \\
4                    & 0.314                        & 0.104                        & \textbf{0.636}               & \textbf{0.654}               & 0.141                        & \textbf{0.496}               \\
6                    & 0.314                        & {\color[HTML]{C00000} 0.111} & {\color[HTML]{C00000} 0.668} & 0.667                        & {\color[HTML]{C00000} 0.198} & 0.477                        \\
8                    & \textbf{0.310}               & 0.088                        & 0.667                        & {\color[HTML]{C00000} 0.668} & 0.197                        & 0.472                        \\ \midrule
\multicolumn{7}{c}{P Value}                                                                                                                                                                                    \\ \midrule
0                    & 0.313                        & \textbf{0.073}               & 0.666                        & \textbf{0.621}               & 0.200                        & 0.481                        \\
0.5                  & 0.316                        & 0.087                        & \textbf{0.643}               & 0.662                        & \textbf{0.147}               & 0.499                        \\
0.6                  & 0.314                        & 0.076                        & 0.666                        & {\color[HTML]{C00000} 0.682} & 0.216                        & 0.507                        \\
0.7                  & 0.312                        & 0.095                        & 0.667                        & 0.647                        & 0.207                        & \textbf{0.534}               \\
0.8                  & {\color[HTML]{C00000} 0.319} & 0.097                        & 0.667                        & 0.675                        & {\color[HTML]{C00000} 0.228} & 0.498                        \\
0.9                  & \textbf{0.310}               & 0.078                        & 0.665                        & 0.667                        & 0.205                        & {\color[HTML]{C00000} 0.462} \\
1                    & 0.315                        & {\color[HTML]{C00000} 0.103} & {\color[HTML]{C00000} 0.668} & 0.663                        & 0.185                        & 0.487                        \\ \midrule
\multicolumn{7}{c}{K Value}                                                                                                                                                                                    \\ \midrule
1                    & 0.313                        & 0.082                        & 0.666                        & \textbf{0.624}               & 0.202                        & 0.481                        \\
20                   & 0.315                        & {\color[HTML]{C00000} 0.115} & 0.664                        & 0.668                        & 0.183                        & {\color[HTML]{C00000} 0.440} \\
30                   & 0.317                        & 0.106                        & {\color[HTML]{C00000} 0.669} & 0.669                        & 0.214                        & \textbf{0.499}               \\
40                   & {\color[HTML]{C00000} 0.322} & 0.091                        & 0.612                        & {\color[HTML]{C00000} 0.674} & {\color[HTML]{C00000} 0.216} & 0.446                        \\
50                   & 0.314                        & 0.090                        & 0.668                        & 0.647                        & 0.197                        & 0.494                        \\
60                   & 0.313                        & 0.106                        & \textbf{0.598}               & 0.657                        & \textbf{0.183}               & 0.467                        \\
70                   & \textbf{0.312}               & \textbf{0.065}               & 0.612                        & 0.646                        & 0.201                        & 0.477                       \\ \bottomrule
\end{tabular}
\caption{Scores for Llama 3 and Llama 3 Instruct on all six psychological metrics for various values of the decoding parameters.}
\label{tab:alignment_table}
\end{table*}
Table \ref{tab:alignment_table} shows the complete results for both Llama 3 variants, with and without instruction finetuning. From the table we can see that the trends are identical among them. Fewer beams and lower P values show better human alignment for both models, with K values showing no consistent trend. However, notably, the model not instruction finetuned appears to show larger improvements in alignment when using decoding methods compared to the instruction finetuned variant. Llama 3 shows a 3.73\%  overall improvement over base when using P=0.5 compared to only 1.40\% for Llama 3 Instruct. 

\subsection{Justification for using cross entropy score}
We compute the four stylistic and psychometric measures as the cross entropy scores between feature vectors of the generated text and the ground truth. These feature vectors are largely all n-dimensional one-hot encoded vectors. Much like multi-class classification tasks where minimizing cross entropy is equivalent to maximizing likelihood, lower cross-entropy for the feature vectors of these four measures indicates higher alignment with the ground truth human dialog.
\end{document}